# A Non-Local Means Approach for Gaussian Noise Removal from Images using a Modified Weighting Kernel

Mojtaba Kazemi
department of electrical, biomedical and mechatronics engineering, Qazvin Islamic Azad
Qazvin, Iran
M.Kazemi@qiau.ac.ir

Ehsan Mohammadi.P
department of electrical, biomedical and mechatronics engineering, Qazvin Islamic Azad
Qazvin, Iran
ehmopi@gmail.com

Parichehr shahidi sadeghi
department of electrical, biomedical and mechatronics engineering, Qazvin Islamic Azad
Qazvin, Iran
Shahidi.parichehr@gmail.com

Mohamad B. Menhaj
Electrical engineering department Amir Kabir university of Technology[*]
Tehran, Iran
Menhaj@aut.ac.ir

*Abstract*— Gaussian noise removal is an interesting area in digital image processing not only to improve the visual quality, but for its impact on other post-processing algorithms like image registration or segmentation. Many presented state-of-the-art denoising methods are based on the self-similarity or patch-based image processing. Specifically, Non-Local Means (NLM) as a patch-based filter has gained increasing attention in recent years. Essentially, this filter tends to obtain the noise-less signal value by computing the Gaussian-weighted Euclidean distance between the patch under-processing and other patches inside the image. However, the NLM filter is sensitive to the outliers (pixels that their intensity values are far away from other pixels) inside the patch, meaning that the pixels with the symmetric locations in the patch are assigned the same weight. This can lead to sub-optimal denoising performance when the destructive nature of noise generates some outliers inside patches. In this paper, we propose a new weighting approach to modify the Gaussian kernel of the NLM filter. Our approach employs the geometric distance between image intensities to come up with new weights for each pixel of a patch, lowering the impact of outliers on the denoising performance. Experiments on a set of standard images and different noise levels show that our proposed method outperforms the other compared denoising filters.

## I. Introduction

Digital images are usually degraded by noise produced during image acquisition, recording, and transmission. The presence of noise can decrease the visual quality of the acquired image and also can negatively affect the performance of other computerized post-processing approaches such as image registration and segmentation [1,2]. Amongst different noise model used for various natural images, the Gaussian noise model is the most prevalent one [3]. Hence, many works have been dedicated to study different characteristics of the Gaussian noise and provide relevant filtering methods to suppress the noise disturbance.

There have been numerous denoising methods to restore the underlying noise-free signal from the given noisy image. A suitable image denoising algorithm should properly remove the noise trace while producing sharp images and preserving the fine details [4,5]. Tomasi *et al*, [6] proposed the bilateral filter, where the authors developed a denoising method based on the SUSAN filter [7] as an extension of the Yaroslavky filter [8]. This method utilized the weighted average of pixels with similar intensity within a local neighborhood to restore the noise-less signal value. However, this filter is not efficient when the noise power is high. Takeda *et al*, [9] proposed an image denoising method using signal-dependent steering kernel regression (SKR) framework, obtaining more robust performance under strong noise condition. Based on the concept of the pixel-based bilateral filter, the well-known Non-Local Means (NLM) method as a patch-based filter approach was proposed in [3], where the concept of locality was extended to the entire image. Kerverran *et al*, [10] developed a denoising method based on the NLM filter that best shows the potential of this method. Using discrete cosine transform and existing data redundancy in the data, BM3D [11] suggests a hybrid approach for grouping similar patches for enoising purposes. In [12,13], a patch-based paradigm for global filtering based on the spectral decomposition was proposed, where each pixel is estimated from all pixels in the image.

A large number of speech and image denoising literature is devoted to transform domain methods (e.g., DCT, Fourier, and wavelet) [14-17]. These filtering methods aim to separate image and noise components in the transform domain, performing denoising on the shrinkage of the transform coefficients. Chang *et al*, [15] employed a spatially adaptive threshold parameter together with the wavelet basis for image denoising purposes. In [16] a transform denoising filter was proposed by modeling the wavelet coefficients of images as mixtures of Gaussians. Luisier *et al*, [17] proposed a denoising method to reduce the mean-squared error (MSE) by wavelet thresholding, reaching a suitable filtering performance.

Dictionary-based learning is another category of image denoising developed based on this notion that similar patches share similar sub-dictionaries; such sub-dictionaries could then be used for better image modeling. Using the dictionary learning basis and sparse representation, Elad *et al*, [18] extended the KSVD approach for image denoising purposes. A nonlocal sparse model (NLSM) was proposed in [19] to improve the performance of the KSVD [20] framework. Note that, the dictionary-based methods essentially provide implicit modeling for natural images. However, Joshi *et al*, [21] focused on the color image denoising by explicitly modeling each pixel as a combination of two colors, where the basis colors are estimated within a local neighborhood. Moreover, in [22], Markov Random Fields (MRFs) as a field of experts (FOE) were applied for image denoising, where the parameters for the model are learned from example images.

[*]His also a part time faculty member of department of electrical, biomedical and mechatronics engineering, Qazvin Islamic Azad University, Menhaj@qiau.ac.ir

In this paper, a modified version of the NLM filter for denoising of the additive Gaussian noise is presented. Essentially, NLM is an efficient filter to deal with Gaussian noise; however, the Gaussian kernel used with this method is sensitive to the outliers (pixels that their intensity values are far away from other pixels) inside each patch, meaning that the pixels with the symmetric locations inside the patch are assigned the same weight. This can lead to sub-optimal denoising performance in case of high noise power, where the destructive nature of noise can generate many outliers in the image. Here, we aim to provide a new weighting scheme to modify the Gaussian kernel of the NLM method. Our approach employs the geometric distance of the image intensities to assign a new weight for each pixel inside the patch. In our experiments, we observed that the geometric distance is a reliable measure in detecting the outliers inside a set of almost similar intensities. Using this new proposed weighting scheme helps lower the impact of outliers on the denoising performance, improving the filtering capabilities.

The rest of this paper is organized as follows: Section II provides the theory of the proposed method. Section III illustrates the experiments and results. Finally, conclusions and some remarks are given in Section IV.

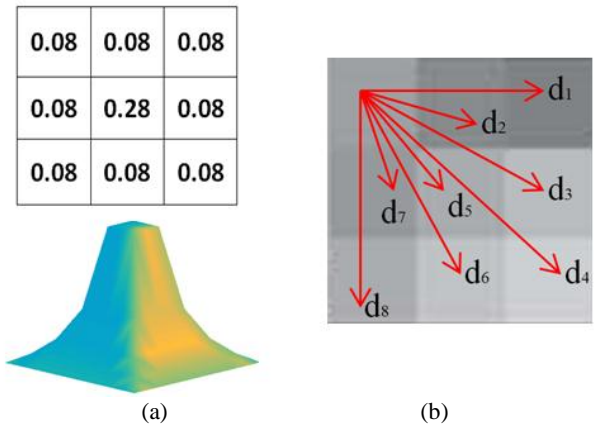

Fig.1. (a) Left side show the Gaussian weighting scheme given by the NLM filter. As seen in the provided patch the weights assigned to the non-central pixels are all equal to 0.08. (b) Right side depicts the proposed geometric distance-based approach that assigned a different weight to each pixel based on the geometric distance of its intensity with respect to other pixels inside the patch.

## II. METHODOLOGY

In this paper, we aim to modify the Gaussian kernel of the NLM filter such that it can suppress the weight of outliers inside each patch under consideration. Hence, the main properties of the NLM filter are briefly reviewed here.

### A. Theory

The NLM filter is proposed based on the Yaroslavsky filter [8], which used to remove noise trace using the average distance between intensities of the image. To find the underlying signal value, the NLM filter evaluates the patch-based similarity between all pixels of the image [3]. Such a patch-based measure is intrinsically more robust than the pixel-based one given by [8], leading to higher denoising performance [3, 23,24]. Given a noisy image $Y$, using the NLM filter the noise-free signal value at a point $p$ is calculated as a weighted average of all the pixels in the image. So, we have:

$$NLM(Y(p)) = \sum_{\forall q \in Y} w(p,q) Y(q) \qquad (1)$$

$$s.t. \quad 0 \leq w(p,q) \leq 1, \quad \sum_{\forall q \in Y} w(p,q) = 1$$

where, $p$ and $q$ respectively represent the pixel under-processing and other pixels in the image. The weights $w(p,q)$ define the similarity between two patches $N_p$ and $N_q$ (i.e., a square neighborhood of size 3 around each pixel) based on a Gaussian kernel function. It is given by:

$$w(p,q) = \frac{1}{\sum_{\forall q} e^{-\frac{d(p,q)}{h^2}}} e^{-\frac{d(p,q)}{h^2}} \qquad (2)$$

$$d(p,q) = G_p \left\| Y(N_p) - Y(N_q) \right\|^2$$

where, $h$ is a constant called smoothing parameter and $G_p$ is the normalized Gaussian weighting function with zero mean and a specific standard deviation $\rho$ (usually set to 1).

The NLM filter has shown remarkable performance in cancelling Gaussian noise. Note that, the main principle behind this approach is employing the existing redundancy of information within the natural images. So, instead of using the pixels inside a local neighborhood, it tends to find the similar pixels throughout the image based on its Gaussian weighting kernel. Then, these weighted pixels are put together to obtain the noise-less underlying signal value.

### B. Modified Kernel Non-Local Means (MK-NLM)

In this paper, we propose a new weighting function to improve the filtering performance of the NLM method. As mentioned earlier, the NLM method assigns same weights to the pixels with the same geometric location inside a patch (see Fig. (1-a)). Essentially, such a weight assignment scheme is not sensitive to the outliers inside the patch, meaning that the outliers are taken as important as other pixels. This issue can be problematic in the presence of heavy noise levels, where the destructive nature of noise may generate many outliers inside the image.

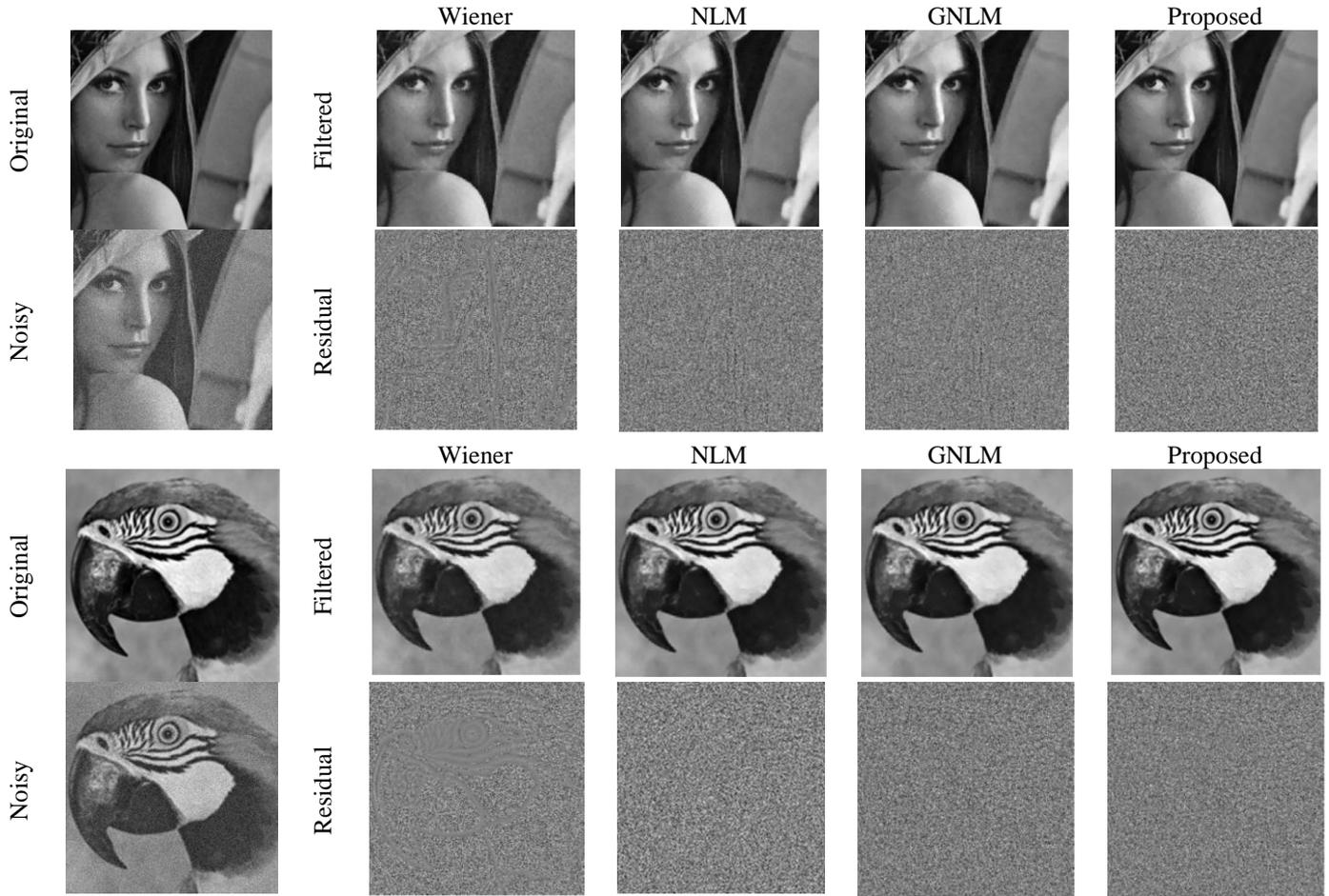

Fig. 2. Visual comparison of filters with 10 % Gaussian noise. In each case, the residual image is shown below the corresponding filtered image. As seen, the filtered image by the proposed method shows remarkable performance in removing noise impact while there is no considerable information and image structure on the residual image.

To address the aforementioned drawback, we propose to modify the weighting scheme given by the NLM filter to reduce the effect of the outliers or singular intensities, decreasing their weights in the calculations. In other words, we consider not only the location of each pixel inside the patch but the self-similarity of pixels as well, decreasing the impact of the outliers in calculating the similarity measure. We employ the geometric distance to isolate the outliers, as it was shown to be a reliable measure in detecting the singular values within a set of almost similar intensities. This is done by calculating the weights between the intensity of each pixel and other existing intensities inside a patch. Fig. (1-b) illustrates our proposed weighting scheme. Thus, our scheme is given by:

$$weight(a_i) = \frac{1}{1+\sqrt{\sum_{j \in N}(a_i - a_j)^2}} \quad 1 \leq i, j \leq N \quad (3)$$

where, *a* represents the intensity of each pixel of a patch with the size equal to *N*. Note that, the calculated geometric distance (i.e., the phrase under the square root) is added by 1 and inverted to calculate the corresponding weight for each pixel. This will ensure that each single weight is normalized in the range of [0,1]. Moreover, it implies that the weights of outliers are set close to zero. Using the weights calculated by Eq. (3), the Gaussian weighting kernel of the NLM filter is modified for each patch as follows:

$$MK\_NLM(p,q) = \frac{G_p \| w(N_p)Y(N_p) - w(N_q)Y(N_q) \|^2}{\sum_{\forall q} w(N_p)w(N_q)}.(4)$$

where, $w(N_p)$ and $w(N_q)$ are respectively the calculated weights for the patch under processing ($N_p$) and other patches ($N_q$). The denominator of Eq. (4) works like a normalization factor to ensure that the new weights are $0 \leq w(N_p), w(N_q) \leq 1$.

III. EXPERIMENTS AND RESULTS

In this section, we compare our proposed method against some existing denoising filters used for Gaussian noise removal purposes.

*A. Quantitative Metrics*

Two well-known quantitative measures are used to evaluate the quality of the filtered image. The first one is Root Mean Squared Error (RMSE) which computes the distance between the ground-truth (*A*) and filtered image (*A'*) of the same size *N* as follows:

TABLE I. QUANTITATIVE COMPARISON OF DIFFERENT DENOISING FILTERS WITH GAUSSIAN NOISE LEVEL OF 5% TO 20% AND DIFFERENT IMAGES. THE BEST VALUE FOR THE RMSE AND SSIM QUALITY MEASURES IS HIGHLIGHTED IN EACH CASE.

| | | | 5% | | 10% | | 15% | | 20% | |
|---|---|---|---|---|---|---|---|---|---|---|
| | | | RMSE | SSIM | RMSE | SSIM | RMSE | SSIM | RMSE | SSIM |
| Lena | 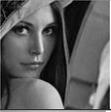 | NLM[3] | 3.058 | 0.951 | 4.250 | 0.901 | 5.421 | 0.847 | 6.756 | 0.785 |
| | | GNLM[12] | **2.630** | **0.967** | 4.181 | 0.912 | 5.283 | 0.875 | 6.458 | 0.830 |
| | | MK-NLM | 2.729 | 0.960 | **4.012** | **0.918** | **5.080** | **0.880** | **6.063** | **0.842** |
| Cameraman | 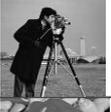 | NLM[3] | 3.650 | 0.958 | 5.483 | 0.907 | 6.508 | 0.843 | 7.764 | 0.786 |
| | | GNLM[12] | 3.615 | 0.952 | 5.120 | 0.919 | 6.320 | 0.865 | 7.557 | 0.821 |
| | | MK-NLM | **3.601** | **0.962** | **4.930** | **0.923** | **6.237** | **0.877** | **7.302** | **0.838** |
| Peppers | 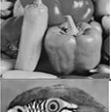 | NLM[3] | 2.511 | 0.965 | 4.205 | 0.926 | 5.610 | 0.875 | 7.004 | 0.820 |
| | | GNLM[12] | **2.483** | **0.975** | **3.992** | 0.941 | 5.465 | 0.895 | 6.950 | 0.861 |
| | | MK-NLM | 2.581 | 0.973 | 4.008 | **0.943** | **5.291** | **0.910** | **6.483** | **0.872** |
| Parrot | 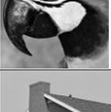 | NLM[3] | 2.810 | 0.969 | 4.386 | 0.928 | 5.802 | 0.872 | 7.182 | 0.814 |
| | | GNLM[12] | 2.852 | 0.958 | 4.222 | 0.933 | 5.608 | 0.892 | 6.885 | 0.849 |
| | | MK-NLM | **2.795** | **0.975** | **4.178** | **0.944** | **5.435** | **0.906** | **6.598** | **0.866** |
| House | 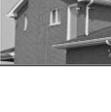 | NLM[3] | 3.010 | 0.9445 | 4.073 | 0.897 | 5.283 | 0.840 | 6.534 | 0.803 |
| | | GNLM[12] | 2.922 | **0.951** | 3.984 | 0.908 | 5.003 | 0.866 | 6.077 | 0.825 |
| | | MK-NLM | **2.712** | 0.949 | **3.840** | **0.912** | **4.845** | **0.878** | **5.845** | **0.837** |

$$RMSE = \sqrt{\frac{1}{N}\sum_{i=1}^{N}\left|A_x - A_x'\right|^2}. \quad (5)$$

We also use the structural similarity (SSIM) index [25] which reveals the perceptual similarity between two images and proved to be more consistent with human visual system. SSIM is given by:

$$SSIM(A, A') = \frac{1}{N}\sum_{x,y=1}^{N}\frac{(2\mu_x\mu_y + c_1)(2\sigma_{xy} + c_2)}{(\mu_x^2 + \mu_y^2 + c_1)(\sigma_x^2 + \sigma_x^2 + c_2)}. \quad (6)$$

where, $\mu_x$ and $\mu_y$ are the local mean values of images $A$ and $A'$, $\sigma_x$ and $\sigma_y$ are the respective standard deviations, $\sigma_{xy}$ is the covariance value and $c_1$ and $c_2$ are two constants.

*B. Filtering Performance*

We evaluate the denoising ability of filters in different noise conditions to study the performance of methods with different noise power. In all cases, the parameters of the denoising methods are set to the best values given by the authors, providing the best performance for each filter. The quantitative results for a wide range of additive Gaussian noise (5-20 percent of the maximum gray level) are reported in Table I. As seen, regardless of the noise power and image in use, the proposed denoising approach returns the best results in most cases. An interesting comparison between methods arises when the noise power is high. As given in Table I, the proposed filter outperforms the other filters when the noise power increases. This shows the higher restoration performance of the proposed filter against other compared filters when the destructive nature of noise is high. For instance, the average RMSE and SSIM measures of the proposed MK-NLM for all employed images are respectively 6.45 and 0.85 as compared to 6.78 and 0.83 given by the GNLM for the noise level 20%. But, the quantitative measures for the same experiments are respectively (2.86 and 0.96) and (2.9 and 0.96) in noise level equal to 5%.

Fig. 2 provides the visual comparison between different denoising methods as well as the proposed approach for 10% of the Gaussian noise level. In addition to the filtered images, the residual images (i.e., the difference between the filtered and the noise-less images) are shown for each filter. Note that, a suitable denoising approach should remove the noise trace as much as possible while maintaining the fine structures of images. Moreover, less information on the residual images is another proper characteristic for the reliable denoising approach. As shown in Fig. 2, the proposed approach shows remarkable capability in removing noise. The results of the GNLM are also comparable to those of the proposed filter. But, on average, it shows weaker performance with the quantitative measures. Furthermore, the residual images obtained by the proposed method mostly include the random characteristic of noise and show no considerable information of the ground-truth.

IV. CONCLUSIONS

This paper proposed a Gaussian noise removal approach based on the well-known nonlocal means (NLM) filter. Intrinsically, the NLM approach provides a patch-based similarity measure based on a Gaussian weighted similarity measure. However, it does not consider the homogeneity of intensities inside the patch under consideration and assigns the same weight to pixels located on the same geometric position. In contrast, we presented a new filtering approach by modifying the weighting scheme of the NLM filter. As a consequence, our proposal takes into account not only the nonlocal similarity among patches, but considers the similarity of pixels intensity inside each patch based on the extracted weighting formulation.

We carried out various experiments on the standard images with different noise levels. The proposed method was compared against several state-of-the-art denoising filters developed to deal with Gaussian noise model. The quantitative results show the superiority of the presented filter in almost all cases, specifically in the high noise levels when the destructive nature of noise destroys the underlying image drastically.

An interesting extension to this work is applying the modified kernel scheme on other kinds of the noise models by customizing the weighting scheme to the noise distribution.